# ROBUST 3D FACE RECOGNITION IN PRESENCE OF POSE AND PARTIAL OCCLUSIONS OR MISSING PARTS

Parama Bagchi[1], Debotosh Bhattacharjee[2] and Mita Nasipuri[3]

[1]Assistant Professor, Department of Computer Science and Engineering, RCC Institute of Information Technology, Beliaghata, Kolkata, India
[2]Associate Professor, Department of Computer Science and Engineering, Jadavpur University, Kolkata, India
[3]Professor, Department of Computer Science and Engineering, Jadavpur University, Kolkata, India

## ABSTRACT

*In this paper, we propose a robust 3D face recognition system which can handle pose as well as occlusions in real world. The system at first takes as input, a 3D range image, simultaneously registers it using ICP(Iterative Closest Point) algorithm. ICP used in this work, registers facial surfaces to a common model by minimizing distances between a probe model and a gallery model. However the performance of ICP relies heavily on the initial conditions. Hence, it is necessary to provide an initial registration, which will be improved iteratively and finally converge to the best alignment possible. Once the faces are registered, the occlusions are automatically extracted by thresholding the depth map values of the 3D image. After the occluded regions are detected, restoration is done by Principal Component Analysis (PCA). The restored images, after the removal of occlusions, are then fed to the recognition system for classification purpose. Features are extracted from the reconstructed non-occluded face images in the form of face normals. The experimental results which were obtained on the occluded facial images from the Bosphorus 3D face database, illustrate that our occlusion compensation scheme has attained a recognition accuracy of 91.30%.*

## KEYWORDS

PCA, REGISTRATION, ICP, RECOGNITION.

## 1. INTRODUCTION

Face is always considered as an important biometric modality in the real world. In comparison to other existing biometric systems like fingerprint recognition, handwriting recognition etc., face biometric can attain a higher performance in security systems. However, the presence of pose changes, occlusion, illumination and expression tend to diminish the performance of a 3D face recognition system. Most importantly 3D facial data are preferred over 2D, because of the following important reasons:- pose changes could be captured well in 3D because of all possible orientations across yaw, pitch and roll. Also, 3D face data are little prone to changes in illumination because, they work with depth values which are unaffected by illumination changes.

The problem occurs when the 3D face is occluded, because occlusion changes the surface information of a 3D face. The detection and simultaneously removal of occlusions, still remains a very challenging area inspite of, some of the work already done in the field of 3D occlusions[9].

Only a few attempts have been made for occlusion detection and removal of those in the field of 3D face have been proven incomplete. For example Park et al. [1] have proposed a method to remove occlusions in the form of only glasses from the 3D image. In [2], the authors have done a similar work on occlusions using Gappy PCA by first registering the faces using ICP but the thresholding technique was done based on sensors.

In [3], the authors have performed a unique classification using a variant of Gappy Principal Component Analysis to remove occlusions and then classification was performed based on region classifiers. In [4], the authors have suggested a method to detect and remove occlusions using a non parametric classification method for occluded face recognition.

The authors in [5] have, at first registered the occluded images using an adaptive based registration scheme, restored occlusions using a masked projection scheme and then performed classification using Fisherface projection. Hassen et. al [6], have used radial curves as a method to remove occlusions from the 3D facial surfaces but, by experimentation they have fixed the initial threshold to detect occlusions.

Our main contribution in the present proposed work is that, we have tried to generalize the thresholding technique for detection of real world occlusions whereas the authors in the above, have all used a predefined threshold for the initial detection of occlusions in real-world face images. Since the preliminary detection of occlusions is very important and working with a fixed threshold cannot be generalized while working on large and different datasets, so our proposed work is a steady step towards generalizing the method of occlusion detection.

The paper is organized as follows: section 2 describes the face detection and normalization module; section 3 reports our significant contributions. Experimental results have been enlisted in Section 4 and hile conclusion and future works have been discussed in section 5.

## 2. PROPOSED SYSTEM

The present proposed system which we have discussed first takes as input a 3D range image, pre-processes it to remove noise and outliers, crops the face image, registers it using ICP and finally the registered image is sent to the occlusion detection section where finally occlusion is removed by PCA method. The module implemented in our present proposed work consists of the following main modules:-

(i) Pre-processing the 3D occluded range image
(ii) Fine registration via Iterative Closest Point (ICP) algorithm, to transform 3D faces into a common coordinate system
(iii) Occlusion detection and removal, where the extraneous objects are found automatically and discarded
(iv) Restoration, to fill in the parts that are labelled as occluded in the previous phase by PCA
(v) Feature extraction and classification by normal points

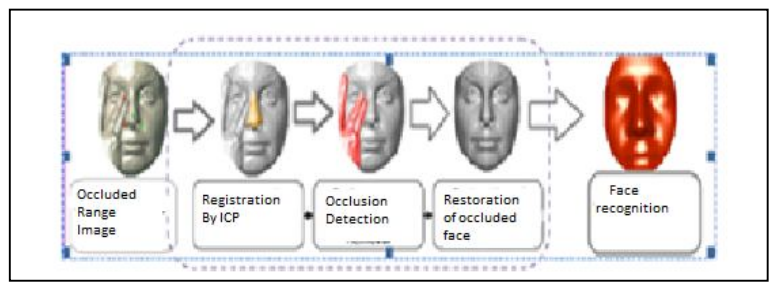

Figure 1. Overall diagram of the proposed method

Figure 1 shows, a more generalized form of the module for occluded region detection, removal and restoration shown by dashed rectangle.

Let us discuss the Steps (i) to (v) one by one:-

**(i) Pre-processing the 3D occluded range image:-** Surface smoothing refers to the fact that noisy spikes and other various deviations are sometimes caused on the 3D face image by noise and other several other factors. So, some types of smoothing techniques are to be applied. In our present technique, we have extended the concept of 2D weighted median filtering technique to 3D face images. The present technique performs filtering of 3D dataset using the weighted median implementation of the mesh median filtering. The surfaces of the 3D artifacts after smoothing, looks as in Figure 2.

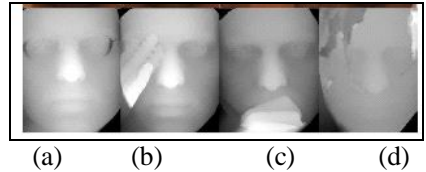

(a)          (b)          (c)          (d)

Figure 2. The four types of occlusions with glasses (a), eyes (b), mouth(c) and hair (d) in the Bosphorus 3D Face Database are shown for a sample individual after smoothing by weighted median filter.

**(ii)Fine registration via Iterative Closest Point (ICP) algorithm[10,11,12], to transform 3D faces into a common coordinate system:-** It is a common phenomenon to be understood that, whenever some occluded objects hide or guard a part of the 3D facial surface, a part of the face's orientation is disturbed to an extent.

For proceeding with the occlusion detection, removal and reconstruction the face must be registered for correct recognition.

The idea of ICP is to refine the position registering the candidate face with a mean face template. We used a customized version of the ICP algorithm aimed to handle the presence of extraneous objects. The algorithm has been inspired by the work presented in [7,8].

ICP algorithm that we have implemented takes as input, a 3D face model in neutral pose and the unregistered occluded 3D range model selected from the 3D Bosphorus database and tries to find a matching criterion in order to find correspondences between the two 3D range models.

The aim of the fine registration step, is to find a rigid transformation that aligns the probe face, P = { $p^1$ …$p^f$ } to the neutral model N={$n^1$ ,……………$n^f$ }. The transformation T can be defined by three rotations around x, y and z axes, $R^x$ ,$R^y$ and $R^z$ and a translation T which is defined by:-

$$T (p^i) = R^x R^y R^z p^i + t; i = 1……..f \qquad (1)$$

where f is the no of points on the 3D surface. Now after registration of the 3D occluded images using the ICP algorithm, the 3D registered images looks as in Figure 3.

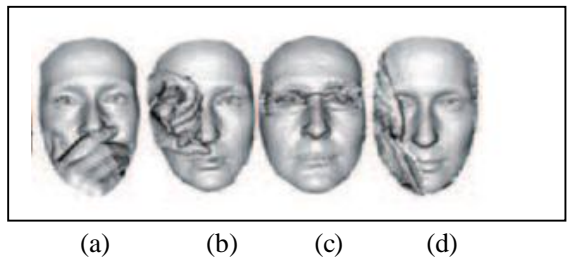

(a)          (b)          (c)          (d)

Figure 3. The four types of occlusions with mouth (a) eye(b) glass (c) hair(d) in the Bosphorus 3D Face Database are shown for a sample individual registration by ICP

**i)** **Performance Analysis of the Registration Technique:-** It is necessary now to mention that, after registration, the mean error rate, needs to be calculated. For this, it is necessary, to calculate the RMS error, between the frontal image as well as the registered image. If v and w are 2 pointclouds then the RMS error is given by:-

$$\mathbf{RMSE\ (v,\ w)} = \sqrt{1/n} \sum_{i=1}^{n}((v_{ix} - w_{ix})^2 + ((v_{iy} - w_{iy})^2 + ((v_{iz} - w_{iz})^2)}$$

In our case, we have found that, the RMSE was found to be between 0.001 to 0.003.

**(iii)Occlusion detection and removal: -** Once registration by ICP is performed for each candidate face, each occluded image is compared with the mean face in order to detect occluding objects. Now, the following condition is checked for each pixel p in the occluded image:-

$$|Y\ (p) - \mu\ (p)| \leqslant T \qquad (2)$$

where Y is the candidate face depth image while $\mu$ is the mean face depth image. But, in most cases the problem lies in selecting the threshold T in Equation 2, which would be valid over multiple databases. There, lies our main contribution.

If the condition in Equation 2 fails, the pixel is invalidated. In this way, large parts of occluding objects can be eliminated. Images with a low number of valid pixels are more difficult to classify because of the lack of information. At this point, images are reconstructed by only the non occluded part of the images.

In order to classify the pixels which are present in the occluded region, a PCA technique has been used to the data sets which are, incomplete. In the present case, after we have detected the occluded part, the resulting image is filled with holes which must be reconstructed for good face recognition. The procedure requires knowledge of which parts of the data are available and which are missing. A set of N patterns {$x_1$, $x_2$... $x_N$}, extracted from a training set of normalized non-occluded faces, and is used to determine the PCA basis in such a way that a generic pattern x can be approximated using a limited number, M, of eigenvectors:-

$$x = \mu + \sum_{i=1}^{M} \alpha_i v_i \qquad (3)$$

where $\mu$ is the mean vector, $v_i$ an eigenvector, and is a coefficient obtained by the inner product between x and v. Now, if there is an incomplete version y of x and suppose that the incomplete vector is M, PCA seeks for an expression similar to Equation 3 for the incomplete pattern y:-

$$y = y' \cong \mu_+ \sum_{i=1}^{M} \beta_i\ v_i \qquad (4)$$

The reconstructed vector y has no gaps since the eigenvectors are complete. Y is hence our final reconstructed non-occluded image. We propose to emphasize the fact that, the reconstruction error is calculated as:-

$$E = ||y' - y|| \qquad (5)$$

This reconstruction error E between the reconstructed image $y'$ and the original image $y$, should be minimized and this is how coefficients of $\beta_i$ should be calculated. The pattern $y'$ can be reconstructed as using Equation 4.

**(iv) Occlusion detection and face restoration:-**

a) Initial detection of occlusion: - Faces are 3D range images projected from the normalized position. Any part of these images that does not look like part of a face, is considered an occlusion [18] (occluding objects may not touch the face). Normally, in most of the previous works described in [1-6], initial detection of occlusions has been very trivial. In the previous works, an initial threshold was formulated on the basis of a trial and error method, just considering the acquisition device. Here, lies our main contribution. It is indeed a very common phenomenon that a single threshold would not work for all databases. So, we have basically tried to generalize the initial occlusion detection procedure in this paper. Let us have a look at Figure 4 where we have generated the depth image corresponding to an occluded image with occlusion in the eyes.

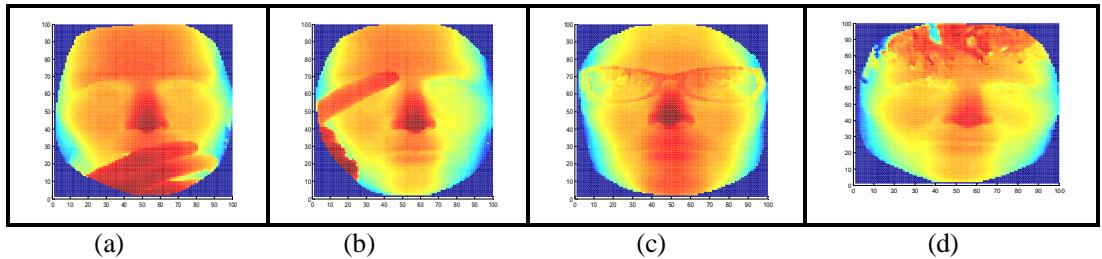

Figure 4. Depth map images of the four types of occlusions with mouth (a) eye(b) glass (c) hair(d) from the Bosphorus 3D Face Database shown for a sample individual corresponding to Figure 3.

After generating the depth map, it could be clearly seen that if the face is sampled at each point on the surface, then the highest intensity values are to be found at each pixel of the occluded portion of the image. But, in the present case the occlusion is to be very vividly present otherwise the present method would not work, since we are working with depth values as a threshold. But it is quite obvious that since the range image is sampled some high intensity values at the non-occluded part of the face would also be present. The procedure for occlusion detection first starts with calculating the difference map between the mean image and the occluded image and storing the difference map in a vector $\vec{e}$. The mean image is basically the non-occluded image, selected from the 3D face database. A snapshot of the difference map image is shown in Figure 5.

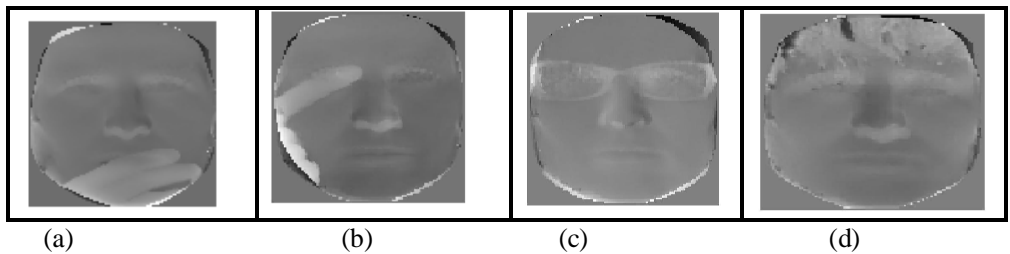

Figure 5. Difference map image images of the four types of occlusions of mouth (a) eye(b) glass (c) hair(d) in the Bosphorus 3D Face Database shown for a sample individual corresponding to Figure 3.

Next, Figure 6 shows a snapshot of the surface of the difference image corresponding to Figure 5(b).

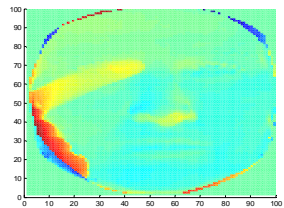

Figure 6. A snapshot of the surface of the difference map image generated for occlusion of the eye from the Bosphorus 3D Face Database shown for a sample individual

The present thresholding technique is described as follows:- After generating the depth map image as is shown in Figure 6, we have calculated the maximum depth value after sampling the image column wise as shown in Figure 7.

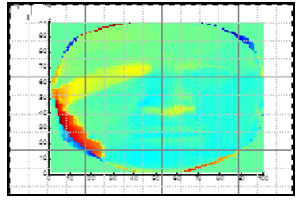

Figure 7. A snapshot of the sampled surface of the difference map image visualized for occlusion of the eye from the Bosphorus 3D Face Database shown for a sample individual

It can be clearly seen that in each column the maximum values are clearly the occluded portions of the image. There could be some outliers, who may occupy the portions of the 3D face, but they are negligible. Now, after thresholding, a preliminary mask M of occlusions can be simply be obtained by the following equation:-

$$M_i = \begin{cases} 1 \ if \ \vec{e} > T \\ 0 \ if \ \vec{e} < T \end{cases}$$

As discussed earlier, $\vec{e}$ is the vector containing the difference map image. T is the threshold, which is obtained, by taking the maximum intensity value of the difference map image sampled at each column of the difference map image. Our approach for obtaining the occlusion mask using the process of thresholding discussed has been enlisted below:-

**Algorithm 1: FindThreshold ($\vec{e}$)**

1.     Assign M = max($\vec{e}$) // M is the variable assigned to find maximum
                                // value of the difference image at each column
2.     for  i  in 1 to r      // r and c is the size of the difference map image $\vec{e}$
3.       for  j in 1 to c
4.         if  $\vec{e} < M$  then Mask(i, j) ← 0
5.         else  Mask(i, j) ← 1
6.       end
7.     end
8.   end

The various occluded masks which we obtained, after thresholding are shown in Figure 8. It is clearly understood from the images that, the white regions represent the areas where occlusion is present, and the black regions represent the non-occluded portions.

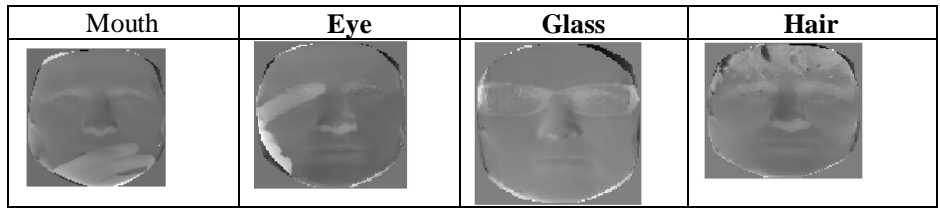

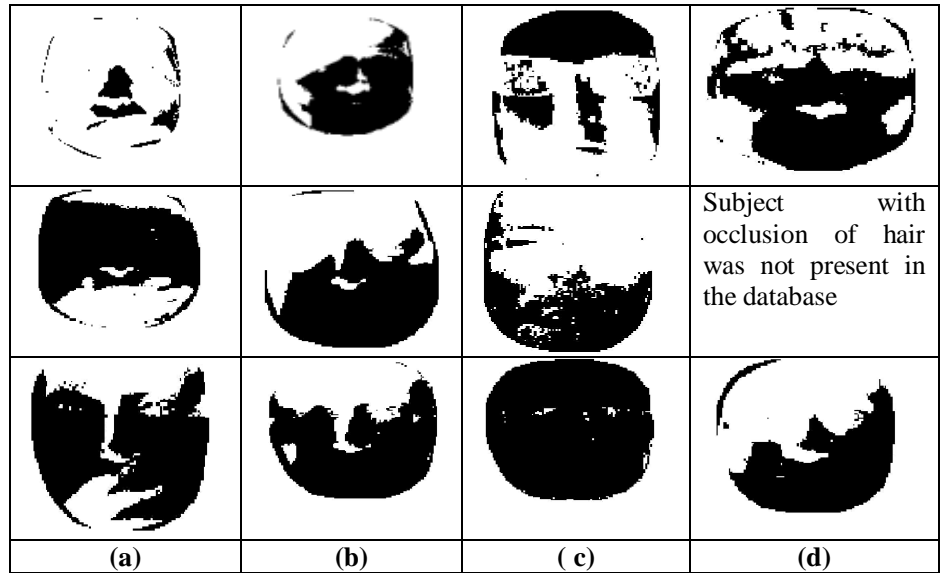

Figure 8 A snapshot of the detection of the occluded region of the surface of the difference map image visualized for occluded faces of sample individuals

Figure 9 shows the edges detected for only the non-occluded regions. The part of the region which is unbounded is the occluded region.

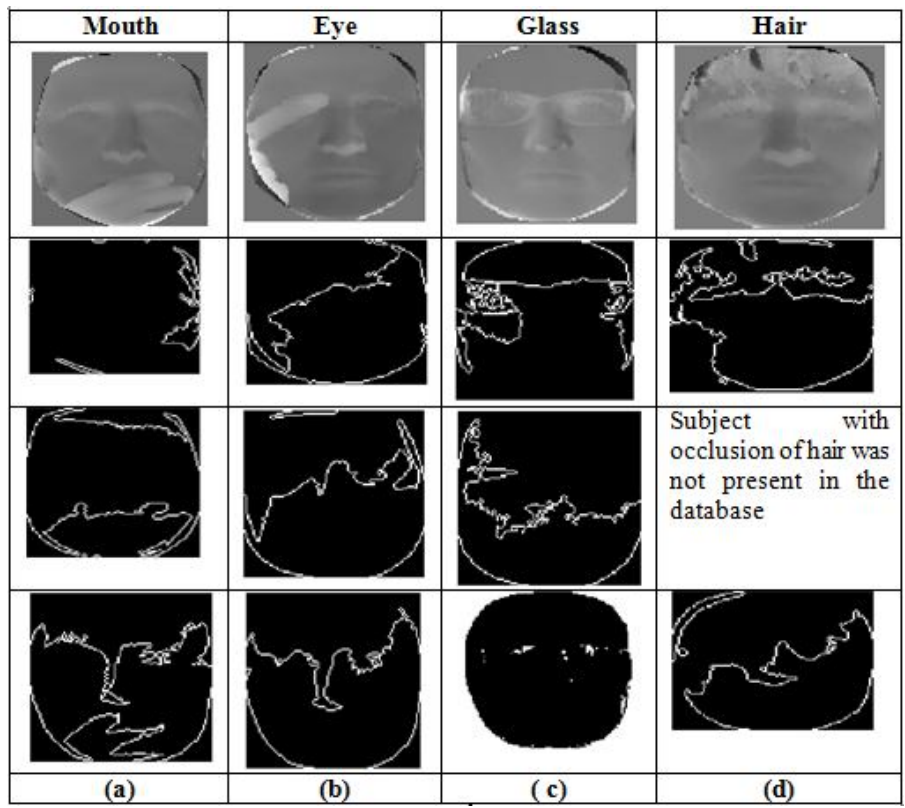

Figure 9. A snapshot of the occluded regions of the surface of the difference map image visualized for occluded faces of sample individuals

In the last row of Figure 9 the occlusion with glass was not well captured in the final segmented image because of occlusion with glasses of that individual was not very vivid and clear.

The algorithm which is used for generating Figure 9 is enlisted as follows:-

---

**Algorithm 2: FindEdges ( $\vec{e}$ )**

---

1.Find the connected components of $\vec{e}$

2.Up_e← Keep only the boundary pixels of $\vec{e}$ and set neighboring pixels to 1

3. Locate all nonzero elements of array Up_e, and returns the linear indices of those elements in vector named ind.

4.Extract the portion of the difference map image corresponding to the coordinates of the indices obtained in Step 3 from the vector ind to get the final boundary of the occluded image.

_______________________________________________________________________________

Figure 9 shows the final face restoration, where we have taken the occluded part from the mean non-occluded face image, and the non-occluded part would remain the same. Now, the image which only has the non-occluded part is filled with holes or in other words, it has incomplete data. Figure 10 below, shows a snapshot of the incomplete data generated for a person with occlusion in the form of hand over mouth.

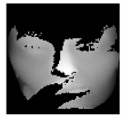

Figure 10. A snapshot of the occluded region of the surface of the difference map image visualized for occlusion of mouth for a face of an individual

At this point, we use the following concept for face restoration:

1.       Using the concept of PCA for face reconstruction

The steps that we have used for face restoration composed of the following steps:-

---

**Algorithm 3: RestoreFace** ( P, N) // P is a vector containing the incomplete face image and N the non- occluded face image.

---

1. Create a vector T, containing a set of vectors, containing non-occluded face images

2.Now create an array of vectors GT, which would contain the vectors of the incomplete image P

3.Calculate the eigenvectors of GT

4.Calculate the coefficients of $\beta_i$, by extracting the most significant eigenvectors of     the incomplete data set.

5.Calculate the coefficients of  $v_i$, by concatenating the incomplete data of vector P

6.The final face image is restored by the following equation:-

$y = y' \cong {}_+ \sum_{i=1}^{M} \beta_i v_i$  // μ  is the mean vector T.

7.Display the final reconstructed image y.

_______________________________________________________________________________

Figure 11(a), (b), (c), (d) shows the final reconstructed image, obtained by PCA.

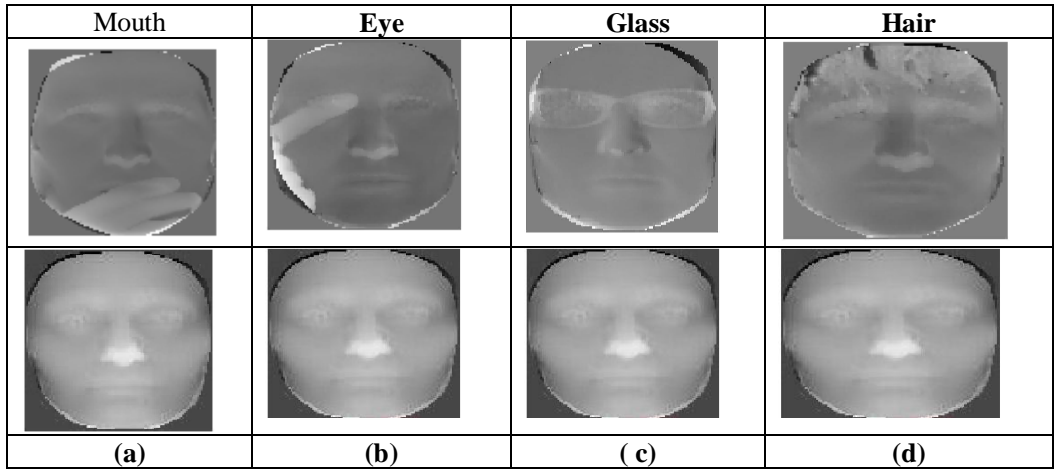

| Mouth | **Eye** | **Glass** | **Hair** |
|:---:|:---:|:---:|:---:|
| **(a)** | **(b)** | **( c )** | **(d)** |

Figure 11 A snapshot of the reconstructed face using PCA method using Algorithm 3 corresponding to the occluded images shown in the first row.

**(v)Feature extraction and classification by normal points:-**The final stage, in the face recognition process is, feature extraction and classification. PCA vector extraction, has been the most common way of feature extraction, but here we have extracted the corresponding normal points, from the restored images as shown in Figure 12 because PCA vector extraction do not give very good results. The reason why, we have extracted the normals, are because of the fact that, normals demonstrate a much better representation. Face normals, are also invariant to expression changes so, they are better candidates, to be used for face recognition purpose. More importantly, in case of any facial structure face normals have been found to be very strong candidates for facial recognition because, they are also very good in establishing correspondences between two different expression changes for e.g. open mouth and close mouth, and also in case of neutral faces.

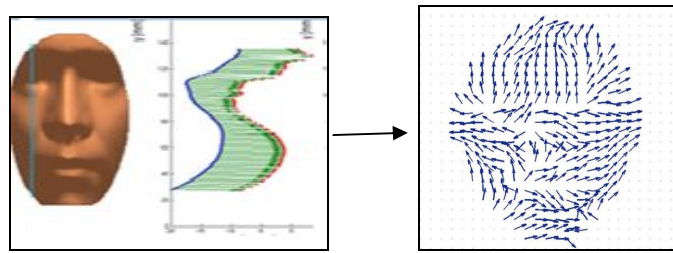

Figure 12. A snapshot of the normals extracted from the non-occluded image.

Now, for each occluded image, i.e. if $x_i$, $y_i$ and $z_i$, where i =1…N, are the 3D coordinates of the final occluded image then, the features are calculated by extracting the normal values of the depth map image, of the final non-occluded 3D image.

## 3. OUR SIGNIFICANT CONTRIBUTION

Now, let us list our significant contributions used in the present proposed work:-

1. In this proposed work, we did not have any masks containing the information for incomplete data. Bosphorus database has no masks containing the information for incomplete data. So our preliminary detection of occlusions using thresholding the depth map, is indeed novel and has not been attempted before.

2. Our feature detection technique for face recognition for the restored images is also novel because in contrast to the already available feature extraction methods, we have used normal features extracted from the depth maps of the restored faces which makes the technique of feature extraction truly novel.

3. PCA is a technique which have been used in many types of 3D facial reconstruction and feature extraction, as well as pose detections and normalization, but here, it has been used on the incomplete masks that we have generated as has been mentioned in Step 1.

4. Finally, we want to justify that, our detection of occlusions and it's simultaneous restoration would be very useful under uncontrolled environments where no initial masks which segregates  the visible parts of the face are available.

## 4. EXPERIMENTAL  RESULTS

In the following section, we provide a comparative performance analysis of our method with other state-of the-art solutions, using only one dataset:- the Bosphorus dataset. The 3D restored, non-occluded images are now fed to the recognition system, to be trained with neural network, for classification purpose. Classification was performed on the extracted normal features, taken from the reconstructed face images using ANN(Artificial Neural network). Bosphorus database composed of 92 subjects, each with 4 types of occlusions namely hair, eye, glasses and mouth, totalling to 368 subjects in the training set. The classification was done primarily in two different ways. We have selected the training and test sets primarily, in two ways:-

Table 1.Table showing subdivision of training and test sets wirh n scans from Bosphorus Database with n=92 classes

| Sl. No. | Test Set | Training Set |
|---|---|---|
| 1. | 1/3rd of the 92 classes | 2/3 rd  of 92 classes |
| 2. | 1/3rd of the 92 classes | 2/3  rd  of  92 classes |

The neural network was trained for 200000 epochs after which the desired recognition rate was reached.

**4.1     3D Face Recognition on the Bosphorus Dataset Under  occlusion:-**  In this section, we have enlisted the performance of  image registration, and the data restoration by PCA technique.

**4.1.1 Comparative study of the performance of registration of the occluded images by ICP:-** Let us discuss first, the registration of occluded images by ICP. The Bosphorus database has 92 subjects each with 4 types of occlusions namely hair, eye, mouth and glasses. After registration of each of these subjects by ICP, each image was successfully registered against the neutral non-occluded model. This is an example of many to one registration, where we are basically registering many occluded models, against a neutral model selected from the Bosphorus database. Table-2 shows a comparative analysis of the results which we obtained over the state-of-art methods for registration only.

Table 2. Comparison of identification results for registration of the present method over other state of art methods with experimentation performed on the Bosphorus database

| | Drira et. al[6] | Alyuz et. al[5] | Alyuz et. al[3] | | Our proposed method |
|---|---|---|---|---|---|
| **Registration technique used** | Recursive ICP | ICP | ICP | | ICP |
| **Occlusions handled** | Occlusion of eye,mouth,hair, glass | Occlusion of eye,mouth,hair, glass | Occlusion of eye,mouth,hair, glass | | Occlusion of eye, outh,hair, glass |
| **Min. Time required for registration** | Not mentioned | Not mentioned | Not mentioned | | 9 sec for each subject |
| **Mean error rate** | Not mentioned | Not mentioned | Not mentioned | | Error rate varied from 0.001 to 0.003 |
| **Performance of Registration** | Not mentioned | Not mentioned | Approaches Used | Performance | Automatic alignment by ICP was 100% successful. The alignment was not done by region classifiers. |
| | | | Automatic nose alignment by ICP | 76.12 | |
| | | | Manual nose AvRM | 76.90 | |
| | | | Manual AvRM | 56.17 | |

Drira et. Al[6], Aluyz[5], and Aluyz[3] also used ICP for registration and all of them worked on the Bosphorus database. The problem was that, the authors did not mention their total time of registering each image from the Bosphorus gallery. Moreover, the registration accuracy was not mentioned by them. In the present proposed work, we have obtained good registration accuracy and all of our images were registered. Figure 13 is a plot showing, the performance of our registration system, that has been enlisted in Table-1 above.

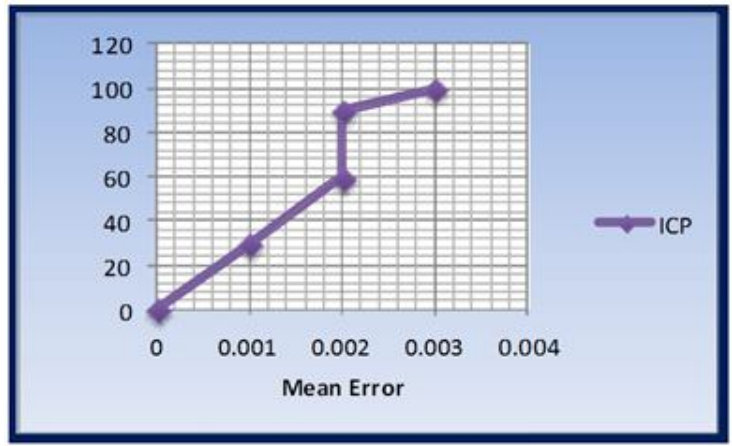

Figure13. A snapshot of the mean error rate after registration by ICP for the present proposed method

**4.1.2 Recognition of the restored images :-** In the case of faces with external occlusion, we first restore them and then apply the recognition procedure. That is, we detect and remove the occluded part, and recover the missing part resulting in a full face that can be compared with a gallery.

The Bosphorus database is suitable for this evaluation, as it contains scans of 60 men and 45 women, 92 subjects in total, in various poses, expressions and in the presence of external occlusions (eyeglasses, hand, hair). The majority of the subjects are aged between 25 and 35. The interesting part is that for each subject there are four scans with occluded parts. These occlusions refer to (i) mouth occlusion by hand, (ii) eyeglasses, (iii) occlusion of the face with hair, and (iv) occlusion of the left eye and forehead regions by hands. Our gallery consisted of the 92 subjects, with each of the occlusions fully restored. So, total we had 368 images which we first randomly divided into:-

The test set as well as the training set were randomly selected. Now, we performed the recognition by extracting features by two different ways and tried to draw a comparison:-

1. Initially, we extracted features using PCA which was the traditional method
2. In the present method, we extracted normal features from the occluded images.

The rank-1 recognition rate, reported for different approaches depending upon the type of occlusion. As these results show the process of restoring occluded parts significantly increases the accuracy of recognition. The rank-1 recognition rate was 66%, when we used approach 1, as is shown in TABLE-3 , by drawing a comparative analysis.

Table 3. (Rank-1 recognition rate) Recognition result comparison of the different methods on the Bosphorus database

|  | Drira et. al[6] | Alyuz et. al[5] | Alyuz et. al[3] | Our proposed method |
|---|---|---|---|---|
| **Feature extraction** | Features in the form of radial curves | Fisherfaces approach | LDA based features | PCA approach |
| **Recognition rate** | 78.63% | 60.63% | 85.04% | 66.66% |

The above methods as in enlisted in Table-3 were occlusion removal without restoration. As is inevitable from the above Table-3, the PCA based approach, works well than Fisherfaces method based feature extraction. But radial curves, works better for when the authors removed the occluded parts and applied the recognition algorithm using the remaining parts. LDA also performs well, but ground truth is involved here.

Now, the rank-2 recognition rate is reported in Table 3, for different approaches depending upon the occlusion after restoration. The rank-2 recognition rate, for the current proposed method was 91.30%, when we used approach 2, as is shown in Table-4 , by drawing a comparative analysis.

Table 4. (Rank-2 recognition rate) Recognition result comparison of the different methods on the Bosphorus database

| | Drira et. al[6] | Alyuz et. al[5] | Alyuz et. al[3] | Our proposed method |
|---|---|---|---|---|
| **Feature extraction** | Features in the form of radial curves | Fisherfaces approach | LDA based features | Feature extraction by normal |
| **Recognition rate** | 87.06% | 93.18% | 94.23% | 91.30% |

It is important to mention that, the work done by Alyuz e al[3], which gave a recognition rate of 94.23%, for the training data set FRGC dataset was used. The authors in [6], had also experimented on Bosporus database, and using occlusion masks to discard the occluded surface patches was beneficial (83.73%), whereas automatically detected occlusion masks provide comparable results to manually labelled occlusion masks (83.99%). Again in [5], the recognition had been performed on the basis of regional classifiers, but we have considered the entire face. That makes our method, truly innovative.

The main reason for performing recognition by PCA is to actually prove that normal features on a 3D surface perform much better than PCA method of feature detection.

## 5. CONCLUSION

3D face recognition has become an emerging biometric technique. However, especially in non cooperative scenarios, occlusion variations complicate the task of identifying subjects from their face images. In this paper, we have presented a fully automatic 3D face recognizer [13,14,15], which is robust to facial occlusions. For the alignment of occluded surfaces, we utilized a ICP registration scheme.
Following the occlusion detection stage, the facial parts detected as occluded are removed to obtain occlusion-free surfaces. Classification is handled on these occlusion-free faces. Here we have incorporated the concept of using normal features for the purpose of classification purposes. We have also seen that, in contrast to intensity based features the present method of normal extraction works much better. Additionally the proposed system can handle small pose variations. The proposed system can handle substantial occlusions and small pose variations.

Similarly, if the face is rotated by more than 30 degrees, it becomes difficult to accomplish the initial alignment.

In our future work, we plan to develop alternative initial alignment techniques. Furthermore, the automatic occlusion detection stage can also be improved: As a future direction, we plan to model occlusions[16,17] better, so that the overall performance of the system can be increased.

## ACKNOWLEDGEMENTS

Authors are thankful to a grant  supported by DeitY, MCIT, Govt. of India, at Department of Computer Science and Engineering, Jadavpur University, India for providing necessary infrastructure to conduct experiments relating to this work.

**Authors**

Parama Bagchi received B.Tech(CSE)and M.Tech(Computer Technology) degree from BCET under WBUT and Jadavpur University in 2005 and 2010 respectively. Her current interests include Image Processing and Face Recognition. She is pursuing her research work in Jadavpur University

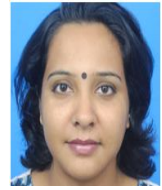

Debotosh Bhattacharjee received the MCSE and Ph.D(Eng.) degrees from Jadavpur University, India, in 1997 and 2004 respectively. He was associated with different institutes in various capacities until March 2007. After that he joined his Alma Mater, Jadavpur University. His research interests pertain to the applications of computational intelligence techniques like Fuzzy logic, Artificial Neural Network, Genetic Algorithm, Rough Set Theory, Cellular Automata etc. in Face Recognition, OCR, and Information Security. He is a life member of Indian Society for Technical Education (ISTE, New Delhi), Indian Unit for Pattern Recognition and Artificial Intelligence (IUPRAI), and senior member of IEEE (USA).

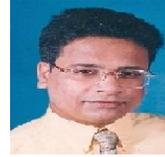

Mita Nasipuri received her B.E.Tel.E., M.E.Tel.E, and Ph.D. (Engg.) degrees from Jadavpur University, in 1979, 1981 and 1990, respectively. Prof. Nasipuri has been a faculty member of J.U since 1987. Her current research interest includes image processing, pattern recognition, and multimedia systems. She is a senior member of the IEEE, U.S.A., Fellow of I.E (India) and W.B.A.S.T, Kolkata, India.

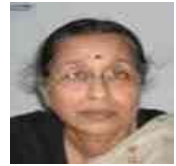